\documentclass[letterpaper]{article} 
\usepackage{aaai2026}  
\usepackage{times}  
\usepackage{helvet}  
\usepackage{courier}  
\usepackage[hyphens]{url}  
\usepackage{graphicx} 
\usepackage{natbib}  
\usepackage{caption} 
\usepackage{algorithm}
\usepackage{algorithmic}
\usepackage{booktabs}
\usepackage{colortbl}
\usepackage{xcolor}
\usepackage{array}
\usepackage{graphicx}
\renewcommand{\arraystretch}{1.1}

\usepackage{float} 
\usepackage{mathtools}
\usepackage{multirow}
\usepackage{graphicx}
\usepackage{booktabs}
\usepackage{enumitem}
\usepackage{algorithm}
\usepackage{algorithmic}
\usepackage{amsmath,amssymb,amsfonts}
\usepackage{textcomp}
\usepackage{subcaption}

\newcommand{\model}{\textsc{$p$LapGNN}}

\usepackage{newfloat}
\usepackage{listings}
\DeclareCaptionStyle{ruled}{labelfont=normalfont,labelsep=colon,strut=off} 
\floatstyle{ruled}
\newfloat{listing}{tb}{lst}{}
\floatname{listing}{Listing}
\title{Enhancing Robustness of Graph Neural Networks through $p$-Laplacian}
\author{Anuj Kumar Sirohi, Subhanu Halder, Kabir Kumar, Sandeep Kumar
}
\affiliations{
    Indian Institute of Technology Delhi, New Delhi, 110016, India\\
    \{aiz218324, eez212381, ee3210741, ksandeep\}@iitd.ac.in
}
\usepackage{bibentry}
\begin{document}
\maketitle
\begin{abstract} 
With the increase of data in day-to-day life, businesses and different stakeholders need to analyze the data for better predictions. Traditionally, relational data has been a source of various insights, but with the increase in computational power and the need to understand deeper relationships between entities, the need to design new techniques has arisen. For this graph data analysis has become an extraordinary tool for understanding the data, which reveals more realistic and flexible modelling of complex relationships. Recently, Graph Neural Networks (GNNs) have shown great promise in various applications, such as social network analysis, recommendation systems, drug discovery, and more. However, many adversarial attacks can happen over the data, whether during training (poisoning attack) or during testing (evasion attack), which can adversely manipulate the desired outcome from the GNN model. Therefore, it is crucial to make the GNNs robust to such attacks. The existing robustness methods are computationally demanding and perform poorly when the intensity of attack increases. This paper presents a computationally efficient framework, namely, \model, based on weighted $ p$-Laplacian for making GNNs robust. Empirical evaluation on real datasets establishes the efficacy and efficiency of the proposed method. 
\end{abstract}

\begin{links}
    {\footnotesize \link{Code}{https://github.com/anujksirohi/pLAPGNN/tree/main}}

\end{links}
\section{Introduction}
In recent years, there has been a significant surge in the need to model and analyze complex interdependencies among entities, for which graphs have emerged as a powerful and expressive representation~\cite{wu2022graph, recommender_application}. Graph-based methods have found widespread applicability across diverse domains, including recommendation systems, protein interaction networks, and the modeling of physical and structural systems. In such representations, edges encode crucial relational information between nodes, often capturing latent or non-obvious properties inherent in the underlying data~\cite{wu2020comprehensive}. Consequently, preserving the integrity of graph structures, particularly in real-time scenarios, is of paramount importance.
\begin{figure}[ht]
\centering
\includegraphics[width=\columnwidth]{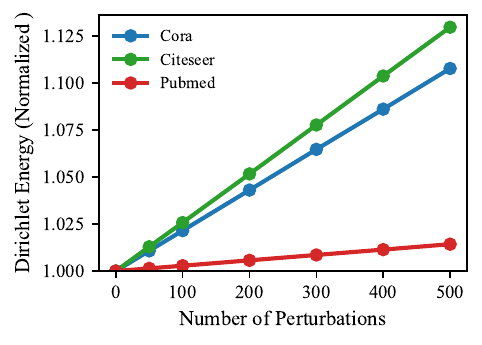}
\caption{Normalized Dirichlet energy under increasing adversarial perturbations on different datasets. A consistent rise indicates a loss of feature smoothness across edges, motivating the need for structure-preserving defenses such as \model.}
\label{fig:dirichlet-relative}
\end{figure}

Graph Neural Networks (GNNs) have recently gained prominence as a robust deep learning paradigm for learning on graph-structured data, achieving notable success across the aforementioned applications~\cite{wu2022graph, wu2020comprehensive, agrawal2024no,sirohi2025hydration}. However, an emerging concern is the vulnerability of these models to adversarial perturbations. Recent studies have demonstrated that carefully crafted attacks can manipulate graph topology, compromising key structural properties and degrading model performance~\cite{sun2022adversarial}. Such perturbations often induce heterophily by creating spurious edges between dissimilar nodes, thereby disrupting the natural homophilic tendencies observed in many real-world graphs. These attacks result in elevated Dirichlet energy, deviating from the smoothness assumptions commonly exploited in graph-based learning (Figure~\ref{fig:dirichlet-relative}). Thus, a central challenge lies in developing mechanisms to restore or preserve the original graph structure in the presence of adversarial manipulations~\cite{9840814}.

Graphs are susceptible to a variety of adversarial attacks that compromise their structural integrity and predictive performance. One common class is \emph{evasion attacks}, which are executed during the testing phase without altering the training data. These attacks can mislead the model into incorrect predictions, exploiting its learned vulnerabilities while preserving the original training distribution. In contrast, \emph{poisoning attacks} manipulate the graph structure during the training phase by perturbing the adjacency matrix or feature space, thereby corrupting the learning process itself~\cite{sun2022adversarial}.

Poisoning attacks can be further categorized as either \emph{targeted} or \emph{non-targeted}. Targeted attacks focus on misclassifying specific nodes of interest, aiming for localized disruption, whereas non-targeted attacks seek to degrade the performance of the model globally by affecting a broad subset of nodes. In this work, we investigate defense strategies against state-of-the-art poisoning attacks, focusing in particular on \textbf{Nettack}~\cite{inproceedings}, a targeted attack that perturbs both features and connections of selected nodes, and \textbf{Metattack}~\cite{Zuegneradv}, a non-targeted meta-learning-based attack that optimizes perturbations to harm the overall model performance.

In recent years, a variety of defense strategies have been proposed to enhance the robustness of Graph Neural Networks (GNNs) against adversarial attacks~\cite{pmlr-v80-dai18b}. One notable example is Jaccard-GCN~\cite{10.1145/3447548.3467416}, which preprocesses the input graph by pruning edges with low Jaccard similarity, thereby mitigating violations of the homophily assumption often exploited by adversarial perturbations. Another approach, GCN-SVD~\cite{10.1145/3336191.3371789}, applies a low-rank approximation to the adjacency matrix to suppress high-frequency noise introduced by attacks, although this method may inadvertently eliminate essential structural information. ProGNN~\cite{graphstructure} addresses this limitation by formulating an optimization problem aimed at recovering a denoised version of the adjacency matrix, leveraging the inherent low-rank and sparse characteristics of graph structures. Additionally, it incorporates a feature smoothing mechanism to down-weight connections between dissimilar nodes. Building upon this, RWLGNN~\cite{9840814} proposes learning the graph Laplacian matrix directly, instead of the adjacency matrix, to exploit its favorable optimization properties and accelerate convergence.

Despite these advances, recent studies have highlighted the advantages of employing the $p$-Laplacian, particularly in settings involving heterophilic graphs, where it demonstrates superior ability to model local inhomogeneities and competitive performance in homophilic scenarios~\cite{slepcev2019analysis, pmlr-v162-fu22e}. Motivated by these findings, we introduce \textbf{\model}, a novel robustness framework that leverages the $p$-Laplacian to denoise adversarially perturbed graphs prior to GNN training. Extensive empirical evaluations on three benchmark datasets confirm the effectiveness of our approach, outperforming existing defense baselines in both targeted and non-targeted attack settings.

\section{Background and Preliminaries}
A graph is formally defined as $\mathcal{G} = (\mathbb{V}, \mathbb{E}, \mathbb{W})$, where $\mathbb{V}$ denotes the set of nodes with $|\mathbb{V}| = n$, such that $\mathbb{V} = \{v_1, v_2, \ldots, v_n\}$. The edge set is given by $\mathbb{E} \subseteq \mathbb{V} \times \mathbb{V}$, and $\mathbb{W} \in \mathbb{R}^{n \times n}$ represents the weighted adjacency matrix. We assume that $\mathcal{G}$ is an undirected, positively weighted graph, i.e., $W_{ij} \geq 0$ for all $i \neq j$, with no self-loops, so that $W_{ii} = 0$ for all $i$. Several matrix representations are commonly used to encode the structure of a graph, such as the adjacency matrix or the (combinatorial) Laplacian matrix, both of which capture connectivity information between nodes and edges \cite{9840814}. In particular, a matrix $\Phi \in \mathbb{R}^{n \times n}$ is called a combinatorial Laplacian if it belongs to the following set \cite{10.5555/3455716.3455738}:  

\begin{align}\label{Lap-set}
\mathcal{S}_{\Phi} = \big\{ \Phi_{ij} = \Phi_{ji} \leq 0 \ \text{for} \ i \neq j, \ \ \Phi_{ii} = -\!\!\sum_{j \neq i} \Phi_{ij} \big\}.
\end{align}
This definition ensures that $\Phi$ is symmetric, diagonally dominant, and positive semidefinite, with zero row and column sums. Consequently, the all-ones vector $\mathbf{1} = [1, 1, \ldots, 1]^{\top}$ lies in its null space, i.e., $\Phi \mathbf{1} = \mathbf{0}$ \cite{10.5555/3455716.3455738}. These properties make the Laplacian a natural choice for developing graph-based learning algorithms.

In addition to the graph topology, we associate each node $v_i \in \mathbb{V}$ with a feature vector $x_i \in \mathbb{R}^d$. Collectively, the features form the matrix $\mathbb{X} = [x_1, x_2, \ldots, x_n]^{\top} \in \mathbb{R}^{n \times d}$, where $d$ denotes the feature dimensionality. For a supervised node classification task, a subset of nodes $\mathbb{V}_L = \{v_1, v_2, \ldots, v_l\} \subset \mathbb{V}$ is provided with ground-truth labels $\mathcal{Y}_L = \{y_1, y_2, \ldots, y_l\}$. The learning objective is to train a model $f_{\theta}$ using $(\mathbb{V}_L, \mathcal{Y}_L)$ such that it generalizes to the remaining unlabeled nodes, i.e.,  
\begin{align}
    f_{\theta} : \mathbb{V} \rightarrow \mathcal{Y},
\end{align}
where $\mathcal{Y}$ is the space of possible class labels. The ultimate goal is to learn $f_{\theta}$ that accurately predicts the labels of nodes in $\mathbb{V} \setminus \mathbb{V}_L$ by leveraging both the node features and the structural information encoded in $\mathcal{G}$. Thus, the training objective of a GNN can be expressed as  
\begin{equation} \label{GNNFormula}
\mathcal{L}_{\mathrm{GNN}}(\theta, \Phi, \mathbb{X}, \mathcal{Y}_L) 
= \sum_{v_i \in \mathbb{V}_L} l\!\left(f_{\theta}(\mathbb{X}, \Phi)_i, \, y_i\right),
\end{equation}
where $l(\cdot,\cdot)$ denotes a loss function (e.g., cross-entropy), and $f_{\theta}(\mathbb{X}, \Phi)_i$ is the predicted label for node $v_i$. However, in the presence of adversarial perturbations, the graph Laplacian $\Phi$ becomes corrupted, yielding a noisy matrix $\Phi_{n}$. Training the GNN directly with $\Phi_{n}$ leads to degraded performance and an unreliable predictor $f_{\theta}$. To address this, our framework proceeds in two stages:  

(i) Noise removal. We first denoise the perturbed Laplacian $\Phi_{n}$ to recover a clean version $\Phi^{*}$ by solving the optimization problem  
\begin{equation}\label{minLnr}
\Phi^{*} = \arg \min_{\Phi} \; \mathcal{L}_{\mathrm{nr}}(\Phi_{n}, \mathbb{X}, \Phi),
\end{equation}
where $\mathcal{L}_{\mathrm{nr}}(\cdot)$ denotes the noise-removal objective.  

(ii) Robust training. Once the clean Laplacian $\Phi^{*}$ is obtained, we train the GNN on $(\Phi^{*}, \mathbb{X})$ to learn a reliable model:  
\begin{equation}\label{minlgnn}
\min_{\theta} \; \mathcal{L}_{\mathrm{GNN}}(\theta, \Phi^{*}, \mathbb{X}, \mathcal{Y}_L).
\end{equation}

In the following section, we discuss in detail the solution strategies for the denoising problem in \eqref{minLnr} and the subsequent robust training formulation in \eqref{minlgnn}.

\section{Algorithm Development}
\begin{figure*}[!htbp]
\centering
\includegraphics[width=\textwidth,keepaspectratio]{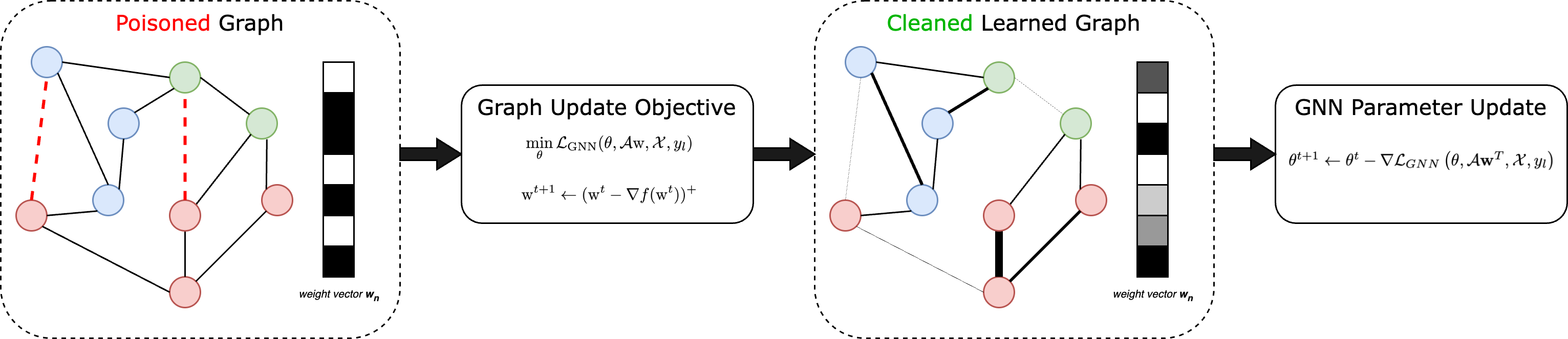}
\caption{Proposed framework. Dashed lines indicate adversarial edges incorporated during the optimization process.}

\end{figure*}
\begin{figure}[!t]
\centering
\resizebox{0.95\columnwidth}{!}{
\begin{minipage}{0.95\columnwidth}
\begin{algorithm}[H]
\caption{\model: MM Denoising and GNN Training}
\label{alg:plapgnn-aaai}
\begin{algorithmic}[1]
\REQUIRE Noisy graph $\mathcal{G}_n=(\Phi_n,\mathbb{X},y_l)$; hyperparameters $\alpha,\beta$; norm $p$; iteration budgets $T_1,T_2$; step sizes $\eta_w$ and $\eta_\theta$
\ENSURE Denoised weights $w^\star$ and trained parameters $\theta$

\STATE \textbf{Precompute:}
\STATE Compute $\delta_{ij} \leftarrow \|x_i-x_j\|_p^p$ for all $i<j$
\STATE Prepare operators $\mathcal{L}$ and $\mathcal{L}^*$, and the index map $k\!\leftrightarrow\!(i,j)$
\STATE Set $L_1 \leftarrow 2n$ and $\eta_w \leftarrow 1/L_1$
\STATE Form vector $\delta \in \mathbb{R}^{n(n-1)/2}$ by stacking $\delta_{ij}$; set $c \leftarrow 2\,\mathcal{L}^*(\Phi_n) - \frac{\beta}{2\alpha}\,\delta$

\STATE Initialize $w^{(0)} \leftarrow w_n$ \quad and \quad $\theta^{(0)} \leftarrow$ random

\STATE \textbf{Stage 1: MM-based Laplacian Denoising}
\FOR{$t=0$ to $T_1-1$}
  \STATE $\nabla f \leftarrow \mathcal{L}^*(\mathcal{L}w^{(t)}) - c$
  \STATE $u \leftarrow w^{(t)} - \eta_w\,\nabla f$
  \STATE $w^{(t+1)} \leftarrow \max(0,\,u)$ \quad /* elementwise projection */
  \STATE \textbf{if} $\displaystyle \frac{\|w^{(t+1)}-w^{(t)}\|_2}{\max(1,\|w^{(t)}\|_2)} < 10^{-4}$ \textbf{ then break}
\ENDFOR
\STATE $w^\star \leftarrow w^{(t+1)}$, \quad $\Phi^\star \leftarrow \mathcal{L}w^\star$
\STATE Build $\mathcal{A}^\star$ from $w^\star$ (symmetric adjacency)

\STATE \textbf{Stage 2: GNN Training}
\FOR{$s=0$ to $T_2-1$}
  \STATE Compute loss $\mathcal{L}_{\mathrm{GNN}}(\theta^{(s)},\mathcal{A}^\star,\mathbb{X},y_l)$
  \STATE $\theta^{(s+1)} \leftarrow \theta^{(s)} - \eta_\theta \,\nabla_\theta \mathcal{L}_{\mathrm{GNN}}(\theta^{(s)},\mathcal{A}^\star,\mathbb{X},y_l)$
  \STATE \textbf{if} early\_stop$(\theta^{(s+1)},\theta^{(s)})$ \textbf{ then break}
\ENDFOR
\STATE \textbf{return} $w^\star,\,\theta^{(s+1)}$
\end{algorithmic}
\end{algorithm}
\end{minipage}
}
\end{figure}
\subsection{$p$-Laplacian Regularization}
The \emph{$p$-Laplacian} operator~\cite{graphstructure} extends the classical graph Laplacian by capturing non-linear smoothness over the graph manifold. When employed as a regularizer in the feature reconstruction or denoising objective, it enables the recovery of more intrinsic structural properties of the original graph.

Let $f: \mathcal{V} \rightarrow \mathbb{R}$ be a real-valued function defined on the vertex set $\mathcal{V}$ of a weighted graph with edge weights $\{w_{ij}\}$. The classical (or $2$-Laplacian) quadratic form is given by~\cite{10.1145/1553374.1553385}:
\begin{equation}
\label{eq:lap2}
\langle f, \Delta_2 f \rangle = \frac{1}{2} \sum_{i,j=1}^{n} w_{ij}\,(f_i - f_j)^2.
\end{equation}
This expression penalizes large differences between function values on adjacent vertices, thereby enforcing local smoothness.

The generalized $p$-Laplacian replaces the squared difference with its $p$-th power, resulting in:
\begin{equation}
\label{eq:lapp}
\langle f, \Delta_p f \rangle = \frac{1}{2} \sum_{i,j=1}^{n} w_{ij}\,|f_i - f_j|^p,
\end{equation}
where $p > 1$. For $p=2$, it reduces to the standard Laplacian, while $p \neq 2$ introduces nonlinearity, allowing for adaptive smoothness control and enhanced robustness to noise and adversarial perturbations in the graph structure.
\subsection{Problem Formulation}

We formulate a joint optimization problem to simultaneously learn robust GNN parameters and recover a clean graph structure from a potentially noisy input graph, following \cite{9840814}. The optimization objective is defined as:
\begin{gather}\label{minThetaPhiC}
    \min_{\theta,\, \Phi^{*} \in S_{\Phi}} 
    \mathcal{L}_{\mathrm{GNN}}\!\left(\theta, \Phi^{*}, \mathbb{X}, y_{l}\right) 
    +  
    \mathcal{L}_{\mathrm{nr}}\!\left(\Phi^{*}, \Phi_{n}, \mathbb{X}\right),
\end{gather}
where the first term corresponds to GNN training using a denoised graph Laplacian matrix $\Phi^{*}$ (to be learned), and the second term enforces structural denoising.

The noise removal objective is given by:
\begin{equation}\label{eq:noise_removal}
    \min_{\mathbf{w} \geq 0} 
    \mathcal{L}_{\mathrm{nr}} 
    = 
    \alpha \, \|\Phi^{*} - \Phi_{n}\|_{F}^{2}
    + 
    \beta \sum_{i,j} \mathbf{w}_{ij} \, \|x_i - x_j\|_{p}^{p},
\end{equation}
where $\Phi_{n}$ denotes the noisy Laplacian, and $\alpha, \beta$ are hyperparameters controlling structural fidelity and denoising strength, respectively.

The first term, $\alpha \|\Phi^{*} - \Phi_{n}\|_{F}^{2}$, constrains the learned Laplacian $\Phi^{*}$ to remain close to the original $\Phi_{n}$, thereby preserving essential topological characteristics of the input graph.  
The second term, $\beta \sum_{i,j} \mathbf{w}_{ij} \|x_i - x_j\|_{p}^{p}$, promotes the suppression of noisy edges between dissimilar node features. The $p$-Laplacian framework enables adaptive control of this term through the parameter \(p\), providing a balance between smoothness and robustness.

The advantages of incorporating the $p$-Laplacian into the denoising objective are summarized as follows:

\begin{itemize}
    \item \textbf{Nonlinearity and Robustness:} For \(p > 2\), the $p$-Laplacian exhibits increased robustness to outliers and large feature discrepancies, effectively pruning edges connecting dissimilar nodes. This contrasts with the Gaussian smoothing behavior inherent in trace-based regularization.
    
    \item \textbf{Sparsity Promotion:} When \(p < 2\), the operator induces sparsity in the learned adjacency by retaining only salient edges while suppressing weak or noisy connections, unlike the uniform smoothing effect of quadratic Laplacian terms.
    
    \item \textbf{Non-Euclidean Flexibility:} The formulation naturally accommodates non-Euclidean feature distances, making it suitable for graphs with irregular or heterogeneous feature spaces.
\end{itemize}

Following prior studies~\cite{10.1145/3336191.3371789, DBLP:journals/corr/abs-1903-01610, graphstructure}, we consider two complementary strategies for optimization:
\begin{itemize}
    \item \textbf{Two-Stage Approach:} The graph is first denoised using the $p$-Laplacian objective, after which GNN parameters $\theta$ are learned on the cleaned structure. This approach is computationally efficient but may yield suboptimal graphs under severe perturbations.
    \item \textbf{Joint Optimization:} Graph denoising and GNN training are performed simultaneously, leading to more robust models in the presence of strong noise, albeit at a higher computational cost.
\end{itemize}

In summary, the proposed joint framework aims to balance efficiency and robustness by leveraging the flexibility of the $p$-Laplacian to adaptively regularize the graph structure during GNN learning.
\subsection{Optimization Procedure}
\textbf{Stage 1: Solving the Noise Removal Objective.}  
The first stage focuses on optimizing the noise removal objective:
\begin{equation}\label{Lnrobj9}
    \min_{\Phi^{*} \in \mathcal{S}_{\Phi}} 
    \mathcal{L}_{\mathrm{nr}}\!\left(\Phi^{*}, \Phi_{n}, \mathbb{X}\right),
\end{equation}
where $\mathcal{S}_{\Phi}$ denotes the set of valid Laplacian matrices satisfying structural constraints. This constitutes a Laplacian-structured constrained optimization problem.

To handle these constraints, we reformulate the problem into a non-negative vector optimization using the Laplacian operator $\mathcal{L}$ defined in~\cite{10.5555/3455716.3455738}. The operator maps a vector of edge weights $w \in \mathbb{R}^{n(n-1)/2}$ to a Laplacian matrix $\mathcal{L}_w \in \mathbb{R}^{n \times n}$ satisfying:
\[
[\mathcal{L}_w]_{ij} = [\mathcal{L}_w]_{ji}, \quad \text{for } i \neq j, 
\quad \text{and} \quad [\mathcal{L}_w]\mathbf{1} = \mathbf{0}.
\]

\textbf{Definition 3.1 (Laplacian Operator).}  
Let $\mathcal{L}: \mathbb{R}^{n(n-1)/2} \to \mathbb{R}^{n \times n}$ be the linear operator mapping $w \mapsto \mathcal{L}_w$, where
\[
[\mathcal{L}_w]_{ij} =
\begin{cases}
    -w_i + d_j, & \text{if } i > j, \\
    [\mathcal{L}_w]_{ji}, & \text{if } i < j, \\
    - \sum_{i \neq j} [\mathcal{L}_w]_{ij}, & \text{if } i = j,
\end{cases}
\]
and $d_j = -j + \frac{(j-1)}{2}(2n-j)$.

\textbf{Definition 3.2 (Adjoint Operator).}  
The adjoint operator $\mathcal{L}^*: \mathbb{R}^{n \times n} \to \mathbb{R}^{n(n-1)/2}$ is defined as:
\begin{equation}
[\mathcal{L}^{*}Y]_k = Y_{ii} - Y_{ij} - Y_{ji} + Y_{jj},
\end{equation}
where $k = i - j + \frac{(j-1)}{2}(2n-j)$.  
These operators satisfy the standard adjoint relation:
\[
\langle \mathcal{L}w, Y \rangle = \langle w, \mathcal{L}^{*}Y \rangle.
\]

Using $\mathcal{L}$, the Laplacian constraint set can be rewritten as:
\begin{equation}\label{Sphi}
    \mathcal{S}_{\Phi} = \{\, \mathcal{L}_w \mid w \geq 0 \,\}.
\end{equation}

Substituting $\Phi^{*} = \mathcal{L}w$ into Eq.~\eqref{Lnrobj9} yields the reformulated objective:
\begin{equation}\label{reformulatedLNR}
    \min_{w \geq 0} 
    \mathcal{L}_{\mathrm{nr}} 
    = 
    \alpha \| \mathcal{L}w - \Phi_n \|_{F}^{2}
    + 
    \beta \sum_{i,j} w_{ij}\|x_i - x_j\|_{p}^{p}.
\end{equation}

This is a non-negative constrained quadratic program of the form:
\begin{equation}
\min_{w \geq 0} 
f(w)
=
\frac{1}{2}\|\mathcal{L}w\|_{F}^{2}
- c^{T}w
+
\frac{\beta}{2\alpha}
\sum_{i,j} w_{ij}\|x_i - x_j\|_{p}^{p},
\end{equation}
where $c = 2(\mathcal{L}^{*}\Phi_n)^{T}$ is precomputed for each dataset under a given perturbation level.

Since the non-negativity constraint $w \geq 0$ precludes a closed-form solution, we adopt a \textit{majorization–minimization} (MM) framework.  
The surrogate function at iteration $t$ is:
\[
g(w \mid w^{(t)}) 
= 
f(w^{(t)}) 
+ (w - w^{(t)})^{T} \nabla f(w^{(t)}) 
+ \frac{L_{1}}{2}\|w - w^{(t)}\|^{2},
\]
where $L_{1} = \|\mathcal{L}\|_{2}^{2} = 2n$ is the Lipschitz constant.

Substituting 
\(
f(w)
=
\frac{1}{2}\|\mathcal{L}w\|_{F}^{2}
- c^{T}w
+
\frac{\beta}{2\alpha}
\sum w_{ij}\|x_i - x_j\|_{p}^{p}
\)
into the surrogate gives:
\[
g(w \mid w^{(t)}) 
= 
\frac{1}{2}w^{T}w 
- a^{T}w,
\quad
\text{where } 
a = w^{(t)} - \frac{1}{L_{1}}\nabla f(w^{(t)}).
\]

Applying the Karush–Kuhn–Tucker (KKT) conditions, the update rule becomes:
\begin{equation}
    w^{(t+1)} = 
    \left( 
        w^{(t)} - \frac{1}{L_{1}}\nabla f(w^{(t)})
    \right)^{+},
\end{equation}
where $(x)^{+} = \max(0, x)$ enforces non-negativity and $t$ indexes the iteration step.  
The gradient is given by:
\begin{equation}
\nabla f(w^{(t)}) = \mathcal{L}^{*}(\mathcal{L}w^{(t)}) - c,
\end{equation}
and we redefine:
\[
c = 2\mathcal{L}(\Phi_n) 
- 
\frac{\beta}{2\alpha}
\sum_{i=1}^{N-1}\sum_{j=i+1}^{N}
\|x_i - x_j\|_{p}^{p}.
\]
This iterative scheme yields the updated weights $w^{(t+1)}$ that progressively denoise the Laplacian structure~\cite{10.5555/3455716.3455738}.

\medskip
\textbf{Stage 2: GNN Parameter Learning.}  
In the second stage, the denoised graph (obtained via $\mathcal{A}w$) is employed for GNN training.  
The model parameters $\theta$ are learned by minimizing:
\begin{equation}
    \min_{\theta} 
    \mathcal{L}_{\mathrm{GNN}}(\theta, \mathcal{A}w, \mathbb{X}, y_{l}),
\end{equation}
where $\mathcal{A}w$ represents the adjacency matrix reconstructed from the optimized edge weights $w$.
\begin{figure*}[!t]
\centering
\begin{subfigure}[t]{0.32\textwidth}
    \centering
    \includegraphics[width=\textwidth]{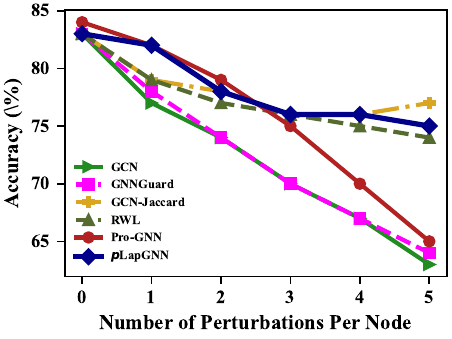}
    \caption{Cora}
    \label{fig:nettack-cora}
\end{subfigure}
\hfill
\begin{subfigure}[t]{0.32\textwidth}
    \centering
    \includegraphics[width=\textwidth]{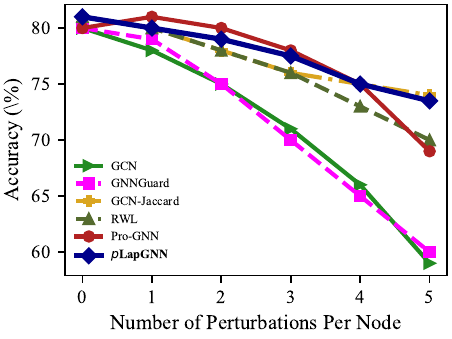}
    \caption{Citeseer}
    \label{fig:nettack-citeseer}
\end{subfigure}
\hfill
\begin{subfigure}[t]{0.32\textwidth}
    \centering
    \includegraphics[width=\textwidth]{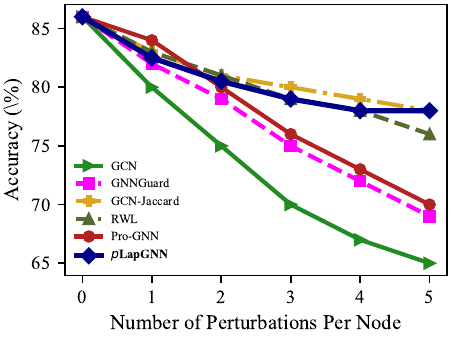}
    \caption{Pubmed}
    \label{fig:nettack-pubmed}
\end{subfigure}

\caption{Performance of various models under Nettack on three benchmark datasets. Our proposed method \textbf{\model} maintains high robustness across all perturbation levels.}
\label{fig:nettack}
\end{figure*}

\begin{table}[t]
\centering
\caption{Summary of datasets, each dataset includes node features, edge connections, and discrete class labels.}
\vspace{0.5em}
\resizebox{\columnwidth}{!}{%
\begin{tabular}{lcccc}
\toprule
\textbf{Dataset} & \textbf{\#Nodes} & \textbf{\#Edges} & \textbf{\#Classes} & \textbf{\#Features} \\
\midrule
Cora     & 2,708 & 5,429 & 7 & 1,433 \\
Citeseer & 3,327 & 4,732 & 6 & 3,703 \\
Pubmed   & 19,717 & 44,338 & 3 & 500 \\
\bottomrule
\end{tabular}
}
\label{tab:dataset_stats}
\end{table}
\begin{table*}[!t]
\centering
\footnotesize
\setlength{\tabcolsep}{5.5pt}
\renewcommand{\arraystretch}{1.1}
\caption{Node classification accuracy (\%) under \textbf{Meta-attack} at various perturbation ratios (PR). 
Results are reported as mean $\pm$ standard deviation over 5 runs. 
Bold indicates the best performance for each setting.}
\label{tab:metaattack_all}
\resizebox{\linewidth}{!}{
\begin{tabular}{l|c|c|c|c|c|c}
\hline
\rowcolor{gray!10}
\textbf{Dataset} & \textbf{PR (\%)} & \textbf{GCN-Jaccard} & \textbf{ProGNN} & \textbf{ProGNN-2} & \textbf{RWLGNN} & \textbf{\model} \\
\hline

\multirow{6}{*}{\textbf{Cora}} 
& 0  & 82.00 $\pm$ 0.20 & 82.98 $\pm$ 0.23 & 73.31 $\pm$ 0.71 & 83.62 $\pm$ 0.82 & \textbf{83.75 $\pm$ 0.71} \\
& 5  & 79.10 $\pm$ 0.21 & \textbf{82.27 $\pm$ 0.45} & 73.70 $\pm$ 1.02 & 78.81 $\pm$ 0.95 & 80.76 $\pm$ 0.40 \\
& 10 & 75.17 $\pm$ 0.42 & 79.03 $\pm$ 0.59 & 73.69 $\pm$ 0.81 & 79.01 $\pm$ 0.84 & \textbf{79.07 $\pm$ 0.61} \\
& 15 & 71.05 $\pm$ 0.30 & 76.40 $\pm$ 1.27 & 75.38 $\pm$ 1.10 & 78.29 $\pm$ 0.86 & \textbf{78.72 $\pm$ 0.57} \\
& 20 & 66.90 $\pm$ 0.40 & 73.32 $\pm$ 1.56 & 73.22 $\pm$ 1.08 & 78.29 $\pm$ 1.05 & \textbf{78.43 $\pm$ 0.70} \\
& 25 & 61.95 $\pm$ 0.50 & 69.72 $\pm$ 1.69 & 70.57 $\pm$ 0.61 & 75.50 $\pm$ 0.72 & \textbf{76.83 $\pm$ 0.50} \\
\hline

\multirow{6}{*}{\textbf{Citeseer}}
& 0  & 71.20 $\pm$ 0.65 & 72.28 $\pm$ 0.69 & 71.20 $\pm$ 0.30 & 70.84 $\pm$ 0.45 & \textbf{73.08 $\pm$ 0.73} \\
& 5  & 70.25 $\pm$ 0.58 & 71.93 $\pm$ 0.57 & 70.10 $\pm$ 0.10 & 71.10 $\pm$ 0.87 & \textbf{72.44 $\pm$ 0.55} \\
& 10 & 69.50 $\pm$ 0.30 & 70.51 $\pm$ 0.75 & 70.08 $\pm$ 0.10 & 70.26 $\pm$ 1.02 & \textbf{71.02 $\pm$ 0.71} \\
& 15 & 65.90 $\pm$ 0.60 & 69.03 $\pm$ 1.11 & 69.02 $\pm$ 0.45 & 68.45 $\pm$ 1.31 & \textbf{69.52 $\pm$ 0.76} \\
& 20 & 59.68 $\pm$ 0.50 & 68.02 $\pm$ 2.28 & 67.20 $\pm$ 0.50 & 67.93 $\pm$ 1.23 & \textbf{68.54 $\pm$ 0.87} \\
& 25 & 59.45 $\pm$ 0.50 & 68.05 $\pm$ 2.78 & 67.20 $\pm$ 0.45 & \textbf{70.40 $\pm$ 0.75} & 69.01 $\pm$ 0.53 \\
\hline

\multirow{6}{*}{\textbf{PubMed}}
& 0  & 87.05 $\pm$ 0.05 & 89.78 $\pm$ 0.32 & 87.15 $\pm$ 0.05 & 89.42 $\pm$ 0.41 & \textbf{90.17 $\pm$ 0.28} \\
& 5  & 86.30 $\pm$ 0.50 & 88.33 $\pm$ 0.45 & 86.70 $\pm$ 0.05 & 88.05 $\pm$ 0.62 & \textbf{89.01 $\pm$ 0.34} \\
& 10 & 85.65 $\pm$ 0.10 & 87.05 $\pm$ 0.63 & 85.85 $\pm$ 0.11 & 87.11 $\pm$ 0.51 & \textbf{87.79 $\pm$ 0.47} \\
& 15 & 84.60 $\pm$ 0.05 & 85.42 $\pm$ 0.76 & 84.44 $\pm$ 0.50 & 85.58 $\pm$ 0.59 & \textbf{86.39 $\pm$ 0.62} \\
& 20 & 83.00 $\pm$ 0.00 & 83.94 $\pm$ 0.98 & 83.00 $\pm$ 0.05 & 84.27 $\pm$ 0.70 & \textbf{85.33 $\pm$ 0.54} \\
& 25 & 82.05 $\pm$ 0.06 & 82.08 $\pm$ 1.13 & 82.00 $\pm$ 0.05 & 82.66 $\pm$ 0.73 & \textbf{83.95 $\pm$ 0.49} \\
\hline
\end{tabular}
}
\vspace{-2mm}
\end{table*}

\section{Experiment}
In this section, we evaluate the effectiveness of the proposed defense mechanism against different types of adversarial attacks. Before presenting the results and observations, we describe the experimental setup.
\subsection{Experimental Settings}
\subsubsection{Datasets}
We validate the performance of the proposed \model~ algorithm on three widely used citation network benchmarks: \textit{Cora}, \textit{Citeseer}, and \textit{PubMed}~\cite{sen2008collective}. These datasets are standard for evaluating robustness and generalization in graph neural networks. Each dataset represents a citation network where nodes denote documents and edges correspond to citation links. The statistics of all datasets are summarized in Table~\ref{tab:dataset_stats}.

\subsubsection{Baselines}
We compare our approach with several state-of-the-art defense methods, including ProGNN~\cite{10.1145/3394486.3403049}, GNNGuard~\cite{NEURIPS2020_690d8398}, GCN-Jaccard~\cite{ijcai2019p669} and RWLGNN~\cite{9840814}. Following the experimental protocol described in~\cite{graphstructure}, all methods employ a two-layer GCN backbone.  

For each dataset, we randomly split the nodes into $80\%$ for training, $10\%$ for validation, and $10\%$ for testing. Hyperparameters are tuned on the validation set to ensure fair comparison across methods.  

\subsection{Defense Performance}
To demonstrate the effectiveness of the proposed \model~framework against state-of-the-art defense methods, we evaluate its node classification performance under various adversarial attack settings. Specifically, we assess robustness against two prominent types of attacks:

\begin{itemize}
    \item \textbf{Nettack:} A targeted attack that perturbs the graph structure around specific nodes to mislead classification. We employ the state-of-the-art targeted attack, Nettack~\cite{inproceedings}, in our experiments.
    \item \textbf{Metattack:} A non-targeted (global) attack that degrades the overall performance of GNNs across the entire graph rather than focusing on specific nodes. We use Meta-Self, a representative variant of the Metattack framework~\cite{Zuegneradv}.
\end{itemize}

We follow the experimental setup of ProGNN~\cite{10.1145/3394486.3403049} for attack generation and data splits. For \textbf{Nettack}, nodes in the test set with a degree greater than $10$ are selected as target nodes, and the number of perturbations per target node varies from $1$ to $5$. For \textbf{Metattack}, the perturbation ratio ranges from $0\%$ to $25\%$ in steps of $5\%$. Additionally, for the \textbf{Random attack}, random edge insertions are applied at rates from $0\%$ to $100\%$ in steps of $20\%$.

The $p$-Laplacian formulation provides several advantages for node classification, particularly in enhancing robustness and convergence. We report node classification accuracies averaged over $10$ runs for Metattack and $5$ runs for Nettack on the Cora, Citeseer, and PubMed datasets under different perturbation levels. Across most attack scenarios, the proposed \model~achieves superior performance compared to baseline methods. Figure~\ref{fig:nettack} shows the performance of \model~under the Nettack attack, where it consistently outperforms or performs on par with the best baseline methods in terms of node classification accuracy. Table~\ref{tab:metaattack_all} summarizes the node classification accuracy of all models under Meta-attacks with varying perturbation ratios. Across datasets, \model~consistently achieves the best or comparable performance. On Cora, it maintains above 76\% accuracy even at 25\% perturbation, outperforming ProGNN and RWLGNN by 3--4\%. For Citeseer, it yields 1--2\% higher accuracy than GCN-Jaccard and ProGNN at moderate perturbation levels. On PubMed, \model~achieves 90.17\% at 0\% and 83.95\% at 25\%, surpassing all baselines while converging faster.
These results confirm that the nonlinear $p$-Laplacian effectively balances robustness, smoothness, and computational efficiency under adversarial conditions. Notably, \model~converges substantially faster requiring only $200$ epochs in the Stage~1 preprocessing phase compared to ProGNN (Joint), which requires $1000$ epochs under Nettack, highlighting its computational efficiency.

\subsection{Parameter Tuning and Implementation Details}
We implemented all attack and defense pipelines using the \texttt{DeepRobust} library, which provides standardized implementations of adversarial attacks and GNN models. Dataset splits follow the ProGNN~\cite{10.1145/3394486.3403049} protocol. For Nettack, nodes in the test set with degrees greater than $10$ were selected as targets, with $1$–$5$ perturbations per node. For Metattack, the perturbation ratio varied from $0\%$ to $25\%$ in increments of $5\%$.  

We used stochastic gradient descent (SGD) as the optimizer for updating weights in the noise-removal stage. The learning rate for the Laplacian optimization was set to $1\times10^{-3}$, while for the GNN model it was set to $1\times10^{-2}$. The optimization to recover the de-noised Laplacian matrix was run for $200$ epochs, and the GNN training was performed for $250$ epochs. Extensive hyperparameter tuning was conducted for $\beta$ in the range $[0.1, 1.5]$ across different perturbation levels. We also varied the $p$ parameter in the $p$-Laplacian to study its influence on performance, while $\alpha$ was fixed at $1$, which consistently yielded the best empirical results.

\subsection{Ablation Study}
To understand the contribution of individual components within our framework, we conduct an ablation study analyzing three key design aspects: the role of the $p$-Laplacian regularizer, the impact of the two-stage optimization, and the influence of the $p$ parameter.

\textbf{Impact of the $p$-Laplacian Regularizer.}  
We compare the full \model~model with a simplified baseline that replaces the nonlinear $p$-Laplacian term with the standard quadratic Laplacian ($p=2$).  
Removing the nonlinear term leads to a consistent reduction in node classification accuracy (by $1.5$–$3\%$) across all datasets and attack types.  
This confirms that the nonlinear operator captures higher-order smoothness and enhances resilience to adversarial edges.

\textbf{Two-Stage vs.\ Joint Optimization.}  
We evaluate a joint optimization variant where graph denoising and GNN parameter learning occur simultaneously.  
While the joint variant achieves comparable performance under low perturbations ($<10\%$), it becomes less stable under higher attack ratios.  
The proposed two-stage strategy maintains higher accuracy and improves training efficiency by approximately $35\%$, validating the benefit of separating denoising from representation learning.

\begin{figure}[h]
\centering
\includegraphics[width=0.99\columnwidth]{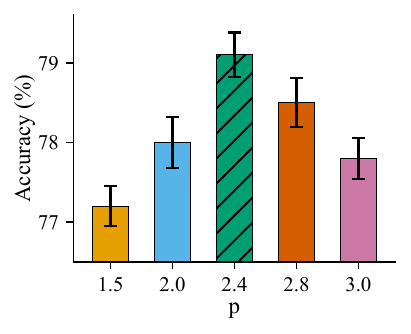}
\caption{Effect of the $p$ parameter on robustness (Cora, 10\% Meta-attack). 
Accuracy peaks at $p=2.4$, indicating a balance between sparsity ($p < 2$) and stability ($p > 2$). Error bars denote standard deviations over five runs.}
\label{fig:p_ablation}
\end{figure}

\textbf{Sensitivity to Parameter $p$.}  
We vary $p$ within the range $[1.5, 3.0]$ to assess its effect on robustness.  
Smaller values ($p<2$) promote sparsity and work well under mild noise, while larger values ($p>2$) yield smoother and more stable graph reconstructions under stronger perturbations.  
Across datasets, $p=2.4$ consistently provides the best trade-off between stability and robustness (see Figure~\ref{fig:p_ablation}).
\begin{table}[t]
\centering
\scriptsize
\setlength{\tabcolsep}{3pt}
\caption{Ablation study on Cora dataset under Metattack ($10\%$ perturbation). Accuracy (\%) averaged over 5 runs.}
\label{tab:ablation}
\begin{tabular}{lcc}
\hline
\textbf{\model~Variant} & \textbf{Description} & \textbf{Accuracy (\%)} \\
\hline
Standard Laplacian ($p=2$) & Without nonlinear regularizer & 77.54 $\pm$ 0.62 \\
Joint Optimization & Denoising and training together & 78.02 $\pm$ 0.48 \\
\model ($p=2.4$) & Two-stage with $p$-Laplacian & \textbf{79.07 $\pm$ 0.61} \\
\hline
\end{tabular}
\end{table}

The ablation results (Table \ref{tab:ablation}) clearly demonstrate that both the nonlinear $p$-Laplacian regularizer and the two-stage optimization contribute significantly to the robustness of the model.  
The adaptive parameter $p$ provides flexibility to handle varying noise levels and adversarial intensities, ensuring stable convergence across diverse graph structures.

\subsection{Convergence, and Complexity Analysis}
The experimental observations align with our theoretical motivation that the $p$-Laplacian introduces an adaptive smoothness constraint capable of adjusting to feature variations among neighboring nodes.  
This adaptive regularization stabilizes gradient propagation during optimization, effectively reducing the influence of adversarial perturbations on graph structure and node embeddings.  
By decoupling the graph denoising and learning stages, $p$LapGNN efficiently filters structural noise while preserving the essential topology, enabling faster convergence and improved generalization across datasets and attack intensities.

From a computational standpoint, the proposed framework introduces minimal overhead and integrates seamlessly with existing GNN architectures such as GCN.  
The overall complexity of \model~remains linear in the size of the graph, i.e., $O(|\mathbb{V}| + |\mathbb{E}|)$, which is comparable to standard message-passing GNNs.  
In contrast, defense methods which rely on repeated singular value decomposition (SVD) operations incur a cubic cost of $O(|\mathbb{V}|^3)$ and significant memory usage.  
The proposed formulation, therefore, achieves robustness without sacrificing scalability, maintaining both computational efficiency and architectural flexibility.

\section{Conclusion}
Graph neural networks are inherently vulnerable to adversarial perturbations, making it essential to recover and preserve the integrity of the underlying graph structure.  
In this work, we introduced \model, a unified framework that employs the nonlinear $p$-Laplacian operator to simultaneously restore the original graph topology and enhance structural awareness within the learning process. This formulation enables the model to learn robust node representations while effectively mitigating the impact of noise and adversarial manipulations. Extensive experiments across multiple benchmark datasets demonstrate that \model~achieves competitive or superior performance compared to existing defense methods, while converging significantly faster. It consistently outperforms existing defense methods under global perturbations (Meta-attack), achieving up to 4\% higher accuracy than the best baseline, and remains within a narrow margin (1--2\%) under targeted attacks (Nettack), all while requiring fewer training epochs and lower computational overhead. For future work, we plan to explore adaptive variants of the $p$-Laplacian and more expressive regularization strategies to further enhance stability, scalability, and generalization under diverse adversarial environments.  
We also aim to extend \model~toward broader graph domains and real-world scenarios, including large-scale heterogeneous networks and dynamic graphs, to assess its effectiveness in practical robustness settings.

\bibliography{aaai2026}
\end{document}